\documentclass[sigconf,nonacm]{acmart} 

\AtBeginDocument{%
  }

\setcopyright{acmlicensed}
\copyrightyear{2025}
\acmYear{2025}
\acmDOI{XXXXXXX.XXXXXXX}
\acmConference[Conference acronym 'XX]{Make sure to enter the correct
  conference title from your rights confirmation email}{June 03--05,
  2025}{Woodstock, NY}
\acmISBN{978-1-4503-XXXX-X/2018/06}

\begin{document}

\title{NEAT: Concept driven Neuron Attribution in LLMs}


\author{Vivek Hruday Kavuri}
\authornotemark[1]
\email{kavuri.hruday@research.iiit.ac.in}
\affiliation{%
  \institution{IIIT Hyderabad}
  \country{India}
}

\author{Gargi Shroff}
\authornote{Both authors contributed equally to this research.}
\email{gargi.shroff@research.iiit.ac.in}
\affiliation{%
  \institution{IIIT Hyderabad}
  \country{India}
}

\author{Rahul Mishra}
\email{rahul.mishra@iiit.ac.in}
\affiliation{%
  \institution{IIIT Hyderabad}
  \country{India}
}

\renewcommand{\shortauthors}{Trovato et al.}

\begin{abstract}
Locating neurons that are responsible for final predictions is important for opening the black-box large language models and understanding the inside mechanisms. Previous studies have tried to find mechanisms that operate at the neuron level but these methods fail to represent a concept and there is also scope for further optimization of compute required. In this paper, with the help of concept vectors, we propose a method for locating significant neurons that are responsible for representing certain concepts and term those neurons as concept neurons. If the number of neurons is n and the number of examples is m, we reduce the number of forward passes required from O(n*m) to just O(n) compared to the previous works and hence optimizing the time and computation required over previous works. We also compare our method with several baselines and previous methods and our results demonstrate better performance than most of the methods and are more optimal when compared to the state-of-the-art method. We, as part of our ablation studies, also try to optimize the search for the concept neurons by involving clustering methods. Finally, we apply our methods to find, turn off the neurons that we find, and analyze its implications in parts of hate speech and bias in LLMs, and we also evaluate our bias part in terms of Indian context. Our methodology, analysis and explanations facilitate understating of neuron-level responsibility for more broader and human-like concepts and also lay a path for future research in this direction of finding concept neurons and intervening them.
\end{abstract}



\maketitle

\section{Introduction}
Large Language Models have demonstrated remarkable capabilities across a wide range of natural language processing tasks. However, the internal mechanisms by which these models represent and utilize abstract concepts remain largely opaque. Understanding which internal components are responsible for specific behaviors is a central goal of \emph{mechanistic interpretability}, as it can enhance transparency, improve controllability, and enable targeted interventions in model behavior.  

\begin{figure}[h]
    \centering
    \includegraphics[width=0.45\textwidth]{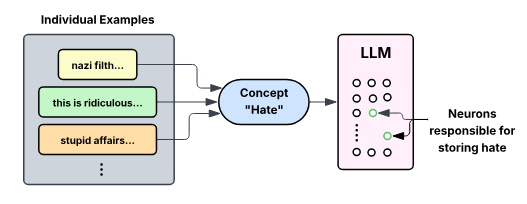}
    \caption{Overall architecture of NEAT: from obtaining the concept vector to identifying and ablating concept-relevant neurons.}
    \label{fig:attention}
\end{figure}

A key challenge in this space is \emph{neuron attribution} — identifying individual neurons or small sets of neurons that significantly influence the model’s predictions. While many attribution methods have been proposed, including causal tracing \cite{vig, meng, zhang2024bestpracticesactivationpatching} and gradient-based approaches \cite{pmlr-v70-sundararajan17a, pmlr-v162-lundstrom22a}, these techniques often require multiple forward and/or backward passes per example, making them computationally expensive at the neuron level. As a result, most prior work has focused on attribution at the \emph{hidden state} level rather than on fine-grained neuron analysis.

Recent advances, such as static neuron importance scoring \cite{yu2024neuronlevelknowledgeattributionlarge}, reduce the computational burden by avoiding multi-pass gradient calculations. While effective for single-instance analysis, these methods do not explicitly address the identification of neurons that encode \emph{abstract, human-interpretable concepts} across diverse contexts. This limitation is particularly important for applications such as bias detection, harmful content mitigation, and concept-level interpretability, where generalization beyond a single example is essential.

In this work, we address this gap by introducing a \emph{concept-vector-based} approach for locating neurons that store information about specific concepts. Our method significantly reduces the computational cost compared to prior works — lowering the forward pass complexity from $O(n \times m)$ to $O(n)$, where $n$ is the number of neurons and $m$ is the number of examples. This is achieved by representing a concept as a single dense vector in the model’s semantic space, then probing all neurons in a single pass to identify those most strongly associated with the concept. We term such neurons \textbf{concept neurons}.

Beyond identification, we investigate \emph{clustering} as an ablation strategy to uncover whether concept-related neurons are organized into semantically coherent groups. This allows us to explore group-level interventions, which can be more efficient and interpretable than modifying neurons individually. We apply our methodology to two real-world settings: \emph{hate speech} and \emph{gender bias}, with the latter including a case study in the Indian socio-cultural context.
Our contributions are as follows:
\begin{itemize}
    \item We propose a novel, computationally efficient method for identifying neurons that represent human-interpretable concepts, reducing forward pass complexity from $O(n \times m)$ to $O(n)$.
    \item We introduce a clustering-based ablation strategy to study group-level neuron interventions. 
    \item We apply our approach to analyze and intervene on hate speech and gender bias in LLMs, providing both quantitative and qualitative evaluations, including a domain-specific bias analysis for the Indian context.
\end{itemize}
Our results demonstrate that the proposed method not only outperforms several baselines in identifying concept-relevant neurons but also provides an interpretable and efficient framework for neuron-level analysis and intervention.

\section{Related Work}
\subsection{Attribution Methods for Transformers}
Identifying the internal components of transformer models that are most influential for a given prediction is a longstanding research challenge. 
Causal tracing approaches \cite{pearl, vig, meng, goldowskydill2023localizingmodelbehaviorpath, zhang2024bestpracticesactivationpatching, wu} and Gradient-based attribution methods \cite{pmlr-v70-sundararajan17a, Kindermans, Miglani, pmlr-v162-lundstrom22a} are widely used to quantify the influence of internal modules in final predictions, but these techniques often require multiple forward and/or backward passes per input making them computationally expensive, especially when applied at the neuron level. Consequently, prior work often focuses on hidden state attribution \cite{meng, geva2023dissectingrecallfactualassociations, stolfo2023mechanisticinterpretationarithmeticreasoning} rather than fine-grained neuron analysis. \\
Recent work by \cite{yu2024neuronlevelknowledgeattributionlarge} proposes a static method for identifying important neurons in LLMs, designed to operate without the heavy computational cost of multi-pass gradient or causal tracing techniques by defining an importance score based on the log-probability increase when a neuron's activation is added to the layer output. While this method is effective for analyzing neuron importance in a single input instance, it doesn't directly address the identification of neurons that store knowledge about broader, abstract concepts across diverse inputs which is crucial for tasks like bias detection and concept-level interpretability, where generalization beyond a single example is required. \\
Another type of studies computes attribution in a single forward pass, relying on proxy saliency measures such as attention weights \cite{vig, Jaunet, Yeh, wang2023labelwordsanchorsinformation, liu2024understandingincontextlearningcontrastive} or feed-forward neuron's coefficient scores. However, these approaches face persistent concerns regarding faithfulness and interpretability \cite{geva2022transformerfeedforwardlayersbuild, lee2024mechanisticunderstandingalignmentalgorithms}, due to lack of standardized evaluation frameworks.

\subsection{Mechanistic Interpretability}
Mechanistic interpretability aims to reverse-engineer the internal computations of language models, often by projecting intermediate activations into the vocabulary space to reveal interpretable patterns \cite{geva2021transformerfeedforwardlayerskeyvalue, dar2023analyzingtransformersembeddingspace, Pal_2023}. Prior studies have identified specialized components such as attention heads \cite{olsson2022incontextlearninginductionheads, wang2022interpretabilitywildcircuitindirect, hanna2023doesgpt2computegreaterthan}, traced bias pathways \cite{vig} and located neurons and hidden states critical for factual knowledge storage \cite{meng, geva2023dissectingrecallfactualassociations}. Work on superposition \cite{elhage2022toymodelssuperposition,  gurnee2023findingneuronshaystackcase} and dictionary learning \cite{bricken2023towards, he2024dictionarylearningimprovespatchfree} explores how features are encoded, with recent focus on neuron level analysis to uncover the mechanisms behind model behavior.

\section{Preliminaries}
First, we introduce the forward pass from inputs to the final predictions - given an input sequence $X = [t_1, t_2, \dots, t_T]$ of $T$ tokens, the model predicts the next token's probability distribution $y$ over vocabulary $V$ of size $B$.  
Each token $t_i$ is embedded via $E \in \mathbb{R}^{B \times d}$ into $h^0_i \in \mathbb{R}^d$, then processed through $L+1$ Transformer layers ($0$ to $L$), each consisting of a multi-head self-attention (MHSA) block and a feed-forward network (FFN):

\begin{equation}
    h^l_i = h^{l-1}_i + A^l_i + F^l_i
\end{equation}

The final output at position $T$ is projected via the unembedding matrix $E_u \in \mathbb{R}^{B \times d}$:
\begin{equation}
    y = \mathrm{softmax}(E_u h^L_T)
\end{equation}

\paragraph{Attention.}  
The attention output is a sum over $H$ heads:
\begin{equation}
    A^l_i = \sum_{j=1}^H \sum_{p=1}^T \alpha^l_{i,j,p} \, W^o_{j,l} \big( W^v_{j,l} h^{l-1}_p \big)
\end{equation}
with attention weights:
\begin{equation}
    \alpha^l_{i,j,p} = \mathrm{softmax}_p \big( W^q_{j,l} h^{l-1}_i \cdot W^k_{j,l} h^{l-1}_p \big)
\end{equation}
where $W^q, W^k, W^v, W^o \in \mathbb{R}^{d \times d/H}$ are the query, key, value and output matrices.

\paragraph{Feed-Forward Network.}  
The FFN output is:
\begin{equation}
    F^l_i = W^{l}_{fc2}\, \sigma\big( W^{l}_{fc1} (h^{l-1}_i + A^l_i) \big)
\end{equation}
where $W^{l}_{fc1} \in \mathbb{R}^{N \times d}$ and $W^{l}_{fc2} \in \mathbb{R}^{d \times N}$.

Following \cite{geva2021transformerfeedforwardlayerskeyvalue}, $F^l_i$ can be expressed as a weighted sum of FFN neurons:
\begin{equation}
    F^l_i = \sum_{k=1}^N m^l_{i,k} \, \mathrm{fc2}^l_k
\end{equation}
where $\mathrm{fc2}^l_k$ are the columns of $W^{l}_{\mathrm{fc2}}$ (called \emph{FFN subvalues}), with coefficient scores:
\begin{equation}
    m^l_{i,k} = \sigma \big( \mathrm{fc1}^l_k \cdot (h^{l-1}_i + A^l_i) \big)
\end{equation}
Here, $\mathrm{fc1}^l_k$ are the \emph{subkeys} (rows of $W^{l}_{\mathrm{fc1}}$).

\paragraph{Neurons.}
From Eqs.~(6)--(7), the $k$-th FFN neuron is defined as the $k$-th subvalue $\mathrm{fc2}^l_k$, weighted by its corresponding coefficient $m^l_{i,k}$ computed from the subkey $\mathrm{fc1}^l_k$.  
Similarly, in Eq.~(3), each attention output can be decomposed into $T \times H$ \emph{position value-output vectors} $W^o_{j,l} \big( W^v_{j,l} h^{l-1}_p \big)$, which can themselves be expressed as weighted sums of \emph{attention neurons}.  
Here, the $k$-th column of $W^o_{j,l}$ is the $k$-th \emph{attention subvalue} and the corresponding $k$-th row of $W^v_{j,l}$ is its \emph{subkey}.  

\section{Methodology}
\begin{figure*}[t] 
    \centering
    \includegraphics[width=\textwidth]{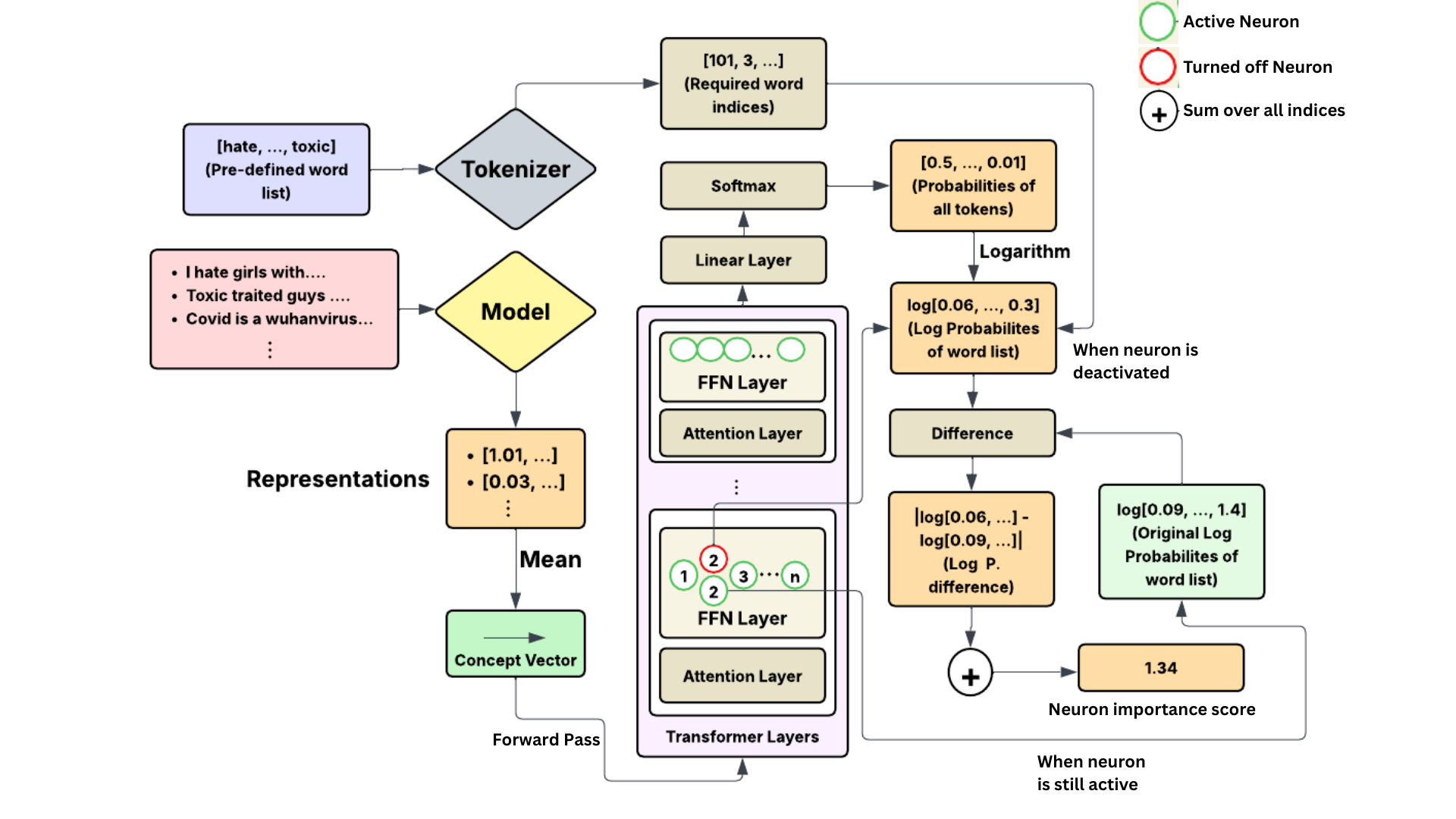} 
    \caption{This image describes our complete overall architecture, from obtaining concept vector to turning a neuron off and seeing it's importance score. Numerals in the diagram are just illustrative and not original values}
    \label{fig:architecture_main}
\end{figure*}
In this section, we first obtain a representational vector for a concept termed as concept vector, given a set of examples for that particular concept. Then, we use the concept vector to locate the important neurons for that particular concept which the concept vector is representing and we term them as \textbf{concept neurons}. Figure ~\ref{fig:architecture_main} shows our overall architecture from obtaining the concept vector from given examples to obtaining the neuron's importance score by turning it off.

\subsection{Obtaining Concept Vector}
\label{sec:concept_vector}
The goal of the concept vector is to represent a concept in the model's semantic space as a vector of dimension as the dimension of model ($D_{\text{model}}$), given $N$ representative examples for the concept.  

We first forward pass all the given $N$ examples representing the concept through the model. For the $i^{\text{th}}$ example, we take its last layer's representation across all tokens, denoted by $\mathbf{h}^{(i)}_t \in \mathbb{R}^{D_{\text{model}}}$ for token $t$, and compute the mean across all $T_i$ tokens:  
\begin{equation}
\mathbf{r}^{(i)} = \frac{1}{T_i} \sum_{t=1}^{T_i} \mathbf{h}^{(i)}_t .
\label{eq:example_repr}
\end{equation}

Then, we average these representations across all $N$ examples to obtain the final concept vector $\mathbf{c} \in \mathbb{R}^{D_{\text{model}}}$:  
\begin{equation}
\mathbf{c} = \frac{1}{N} \sum_{i=1}^N \mathbf{r}^{(i)} .
\label{eq:concept_vector}
\end{equation}

\subsection{Locating Concept Neurons}
To locate the concept neurons, we begin by defining a list of $K$ target words from the model's vocabulary, $\mathcal{W} = \{ w_1, w_2, \dots, w_K \}$, which are representative of the concept of interest. For example, in the case of a ``hate'' concept, this set may consist of hate-related terms. We obtain the token IDs for these words from the tokenizer and denote them as $\mathcal{T} = \{ t_1, t_2, \dots, t_K \}$.

Given the concept vector $\mathbf{c}$ from Equation~\ref{eq:concept_vector}, we feed it to the model as an input embedding to obtain intermediate representations at every transformer layer. 

We measure the importance of a neuron by selectively \emph{deactivating} it, i.e., zeroing out its contribution in the feed-forward projection. For a given layer $l$ with projection matrix $W_{\text{proj}}^{(l)} \in \mathbb{R}^{D_{\text{model}} \times D_{\text{ffn}}}$, the $j^{\text{th}}$ neuron corresponds to the $j^{\text{th}}$ column vector $\mathbf{w}^{(l)}_j$. Deactivating this neuron is equivalent to:
\begin{equation}
\tilde{W}_{\text{proj}}^{(l)}[:, j] = \mathbf{0},
\label{eq:zero_neuron}
\end{equation}
where $\tilde{W}_{\text{proj}}^{(l)}$ denotes the modified projection matrix.

Given the residual stream vector before the projection step, $\mathbf{r}^{(l)} \in \mathbb{R}^{D_{\text{model}}}$, we compute the original feed-forward output as:
\begin{equation}
\mathbf{o}_{\text{orig}}^{(l)} = \mathbf{r}^{(l)} + W_{\text{proj}}^{(l)} \, \sigma \!\left( W_{\text{fc}}^{(l)} \mathbf{r}^{(l)} \right),
\label{eq:ffn_orig}
\end{equation}
and the modified output after neuron deactivation as:
\begin{equation}
\mathbf{o}_{\text{mod}}^{(l,j)} = \mathbf{r}^{(l)} + \tilde{W}_{\text{proj}}^{(l)} \, \sigma \!\left( W_{\text{fc}}^{(l)} \mathbf{r}^{(l)} \right),
\label{eq:ffn_mod}
\end{equation}
where $W_{\text{fc}}^{(l)}$ is the first feed-forward projection and $\sigma(\cdot)$ denotes the activation function.

We then pass $\mathbf{o}_{\text{orig}}^{(l)}$ and $\mathbf{o}_{\text{mod}}^{(l,j)}$ through the final layer normalization $\mathrm{LN}_f$ and the language modeling head $W_{\text{lm}}$ to obtain the logits:
\begin{align}
\mathbf{z}_{\text{orig}} &= W_{\text{lm}} \, \mathrm{LN}_f \!\left( \mathbf{o}_{\text{orig}}^{(l)} \right), \\
\mathbf{z}_{\text{mod}}^{(l,j)} &= W_{\text{lm}} \, \mathrm{LN}_f \!\left( \mathbf{o}_{\text{mod}}^{(l,j)} \right).
\end{align}
The corresponding probability distributions are then obtained by applying a softmax:
\begin{align}
p_{\text{orig}}(t) &= \frac{\exp(z_{\text{orig},t})}{\sum_{u} \exp(z_{\text{orig},u})}, \\
p_{\text{mod}}^{(l,j)}(t) &= \frac{\exp(z_{\text{mod},t}^{(l,j)})}{\sum_{u} \exp(z_{\text{mod},u}^{(l,j)})}.
\end{align}

Finally, the \emph{target effect score} for neuron $(l, j)$ is computed as the total absolute log-probability change over the target token set $\mathcal{T}$:
\begin{equation}
E_{l,j} = \sum_{t \in \mathcal{T}} \left| \log p_{\text{mod}}^{(l,j)}(t) - \log p_{\text{orig}}(t) \right|.
\label{eq:target_effect}
\end{equation}

We rank all neurons by $E_{l,j}$ in descending order and select the top $M$ neurons with the largest values as the \textbf{concept neurons} for the given concept. The output of this step is a set:
\begin{equation}
\mathcal{N}_{\text{concept}} = \operatorname*{arg\,topM}_{l,j} \; E_{l,j}.
\label{eq:concept_neurons}
\end{equation}

These neurons are the ones whose removal causes the greatest change in the likelihood of concept-representative words, indicating that they are most strongly associated with encoding the given concept.

\subsection{Extension to Bias}
\label{sec:extension_to_bias}
We extend the methodology from the previous subsection to identify neurons specifically associated with \emph{gender-related stereotypical bias}. We treat \emph{male-stereotype} and \emph{female-stereotype} as two distinct concepts and obtain their respective concept vectors.

Let $\mathcal{S}_{\text{M}} = \{ s^{(1)}_{\text{M}}, \dots, s^{(L)}_{\text{M}} \}$ denote $L$ \emph{male-stereotyped} sentences, and $\mathcal{S}_{\text{F}} = \{ s^{(1)}_{\text{F}}, \dots, s^{(L)}_{\text{F}} \}$ denote $L$ \emph{female-stereotyped} sentences. Using Equations~\ref{eq:example_repr}--\ref{eq:concept_vector}, we obtain:  
\begin{align}
\mathbf{c}_{\text{M}} &= \frac{1}{L} \sum_{i=1}^L \mathbf{r}_{\text{M}}^{(i)}, \\
\mathbf{c}_{\text{F}} &= \frac{1}{L} \sum_{i=1}^L \mathbf{r}_{\text{F}}^{(i)},
\end{align}
where $\mathbf{c}_{\text{M}}, \mathbf{c}_{\text{F}} \in \mathbb{R}^{D_{\text{model}}}$ are the male- and female-stereotype concept vectors respectively.

We define two fixed sets of gendered words:  
\begin{align}
\mathcal{T}_{\text{M}} &= \{ t_{\text{he}}, t_{\text{father}}, t_{\text{king}}, \dots \}, \\
\mathcal{T}_{\text{F}} &= \{ t_{\text{she}}, t_{\text{mother}}, t_{\text{queen}}, \dots \},
\end{align}
where each element is a token ID obtained from the tokenizer.

\paragraph{Male-stereotypical neurons}  
We forward pass $\mathbf{c}_{\text{M}}$ through the model and compute the \emph{target effect score difference} for each neuron $(l,j)$ as:
\begin{equation}
E^{\text{M-bias}}_{l,j} = \sum_{t \in \mathcal{T}_{\text{M}}} \left| \Delta \log p^{(l,j)}(t) \right|
\;-\;
\sum_{t \in \mathcal{T}_{\text{F}}} \left| \Delta \log p^{(l,j)}(t) \right|,
\label{eq:male_bias_score}
\end{equation}
where $\Delta \log p^{(l,j)}(t) = \log p_{\text{mod}}^{(l,j)}(t) - \log p_{\text{orig}}(t)$, and $p_{\text{orig}}, p_{\text{mod}}^{(l,j)}$ are defined as in the previous subsection.  
The neurons with the largest positive values of $E^{\text{M-bias}}_{l,j}$ are identified as \textbf{male-stereotypical neurons}:
\begin{equation}
\mathcal{N}_{\text{M-bias}} = \operatorname*{arg\,topM}_{l,j} \; E^{\text{M-bias}}_{l,j}.
\end{equation}

\paragraph{Female-stereotypical neurons}  
Similarly, we forward pass $\mathbf{c}_{\text{F}}$ and compute:
\begin{equation}
E^{\text{F-bias}}_{l,j} = \sum_{t \in \mathcal{T}_{\text{F}}} \left| \Delta \log p^{(l,j)}(t) \right|
\;-\;
\sum_{t \in \mathcal{T}_{\text{M}}} \left| \Delta \log p^{(l,j)}(t) \right|,
\label{eq:female_bias_score}
\end{equation}
The neurons with the largest positive $E^{\text{F-bias}}_{l,j}$ are identified as \textbf{female-stereotypical neurons}:
\begin{equation}
\mathcal{N}_{\text{F-bias}} = \operatorname*{arg\,topM}_{l,j} \; E^{\text{F-bias}}_{l,j}.
\end{equation}

This formulation ensures that neurons are selected based on their asymmetric effect on gendered tokens, thereby isolating those that are more strongly associated with one gender’s stereotypical representation than the other.
\section{Experiments}
In this section we describe the models and datasets we have use to apply the method described in the methodology section and also discuss the ablation studies that we have done.
\subsection{Models and Datasets}
\paragraph{Model}  
For all experiments, we use the \texttt{gpt2-large} model , which is a transformer-based causal language model with $36$ layers, a hidden dimension $D_{\text{model}} = 1280$, and $774$ million parameters. 

\paragraph{Datasets}  
We apply our methodology on two distinct settings:  

\textbf{Hate Speech Setting:}  
    To obtain representative examples for the ``hate'' concept, we use the \emph{Wiki Hate} dataset \cite{wulczyn2017ex}, which contains Wikipedia comments annotated for hatefulness. We select a subset of 2000 sentences explicitly labeled as hateful to construct the concept vector.
    
\textbf{Bias Setting:}  
    To study gender-related stereotypical bias, we use the \emph{Bias in Bios} dataset \cite{10.1145/3287560.3287572}, which consists of biographies labeled with the subject's gender and profession. We also label the each of the profession as whether it is male stereotypical profession or female stereotypical profession manually. We then extract two balanced sets, each of 2000 of sentences: one containing \emph{male-stereotyped} examples and the other containing \emph{female-stereotyped} examples. These are used to form the respective male and female concept vectors, as described in Section~\ref{sec:extension_to_bias}.

In both settings, text is preprocessed through the default GPT2 tokenizer. After obtaining the neurons using the methodology described in the previous section, we turn of the top 10 neurons in hate speech setting and evaluate. For bias setting, we incrementally turn off neurons among the top neurons obtained and evaluate. Exact evaluation pipeline will be discussed in the further sections.

\subsection{Ablation Study — Clustering}  
In addition to identifying concept neurons individually, we investigate whether neurons responsible for encoding a concept are organized into coherent \emph{clusters} in the model's parameter space. The motivation for this experiment is twofold:  
(1) If concept-related neurons form tight clusters, group interventions might be a more efficient and robust way to modify a model’s behavior than targeting neurons individually.  
(2) Clustering may reveal distributed representations, where individual neurons contribute partially to a concept but become significant when considered jointly.  

\paragraph{Clustering Procedure.}  
For a given concept vector $\mathbf{c}$ (obtained as in Section~\ref{sec:concept_vector}), we perform a forward pass through the model and record the \emph{coefficient scores} $m^l_{i,k}$ (Eq.~\ref{eq:example_repr}--\ref{eq:concept_vector}) for all neurons across all layers, where $i$ indexes the concept vector input position and $(l,k)$ identifies the neuron. Since we feed the concept vector as a single embedding, the coefficient score for neuron $(l,k)$ is:  
\[
m^{(l,k)}_{\text{concept}} = \sigma \big( \mathrm{fc1}^l_k \cdot \mathbf{c} \big) .
\]  
We then collect all such scores into a feature matrix $\mathbf{M} \in \mathbb{R}^{N_{\text{total}} \times 1}$, where $N_{\text{total}}$ is the total number of FFN neurons in the model ($L \times N_{\text{ffn}}$).  

We apply \emph{k-means clustering} on these coefficient scores to partition neurons into $k$ groups:
\[
\mathcal{C}_1, \mathcal{C}_2, \dots, \mathcal{C}_k, \quad \bigcup_{u=1}^k \mathcal{C}_u = \{ 1, \dots, N_{\text{total}} \}.
\]  
Clusters are ranked according to their mean target effect score (Eq.~\ref{eq:target_effect}) across neurons in the cluster:
\[
E_{\text{cluster}}(u) = \frac{1}{|\mathcal{C}_u|} \sum_{(l,j) \in \mathcal{C}_u} E_{l,j}.
\]
The cluster with the highest $E_{\text{cluster}}(u)$ is deemed the \emph{most concept-relevant cluster}.  

\paragraph{Cluster-based Intervention.}  
To test whether concept-related knowledge is localized, we perform \emph{cluster-level interventions}: instead of deactivating individual top-$M$ neurons, we simultaneously deactivate all neurons in the top-ranked cluster by zeroing their contributions as in Eq.~\ref{eq:zero_neuron}.  

\paragraph{Varying $k$.}  
We repeat the above process for different values of $k$ to analyze how cluster granularity affects performance. The results and analysis for the same is provided in the Appendix \ref{sec:appendix}  



\section{Results}
In this section we start by discussing our evaluation pipeline in which we discuss the data used for evaluation and the metrics we use. Then we end the section by presenting and discussing the results.
\subsection{Evaluation Pipeline}
\subsubsection{Hate Speech}
For the hate speech evaluation, we use the \emph{Wiki Hate} dataset, from which we sample a subset of 2000 sentences containing explicit or implicit hate-related content. Then, we get the important neurons for the hate concept from the examples we have selected.

We evaluate the model’s ability to predict the last token in each sentence after deactivating the selected neurons. We first evaluate by forward passing without neuron deactivation to establish a baseline. Then, the top-$10$ neurons identified from the different methods are deactivated and tested on the top 10 sentences from the sampled 2000 sentences for the metrics as described below:

\begin{enumerate}
    \item \textbf{Mean Reciprocal Rank (MRR)} of the true last token:  
    Let $r_i$ denote the rank position of the correct last token for the $i$-th sentence in the predicted probability distribution.  
    The reciprocal rank for sentence $i$ is given by:
    \[
        \mathrm{RR}_i = \frac{1}{r_i}
    \]
    The MRR over $N$ sentences is:
    \[
        \mathrm{MRR} = \frac{1}{N} \sum_{i=1}^{N} \mathrm{RR}_i
    \]

    \item \textbf{Probability} assigned to the true last token:  
    Let $y_i$ denote the index of the correct last token for sentence $i$ and $\mathbf{p}_i$ denote the predicted probability distribution (after softmax). The probability is:
    \[
        P_i = p_i(y_i)
    \]
    where $p_i(y_i)$ is the predicted probability of the correct token.

    \item \textbf{Log-Probability} of the true last token:  
    We compute:
    \[
        \log P_i = \log\left( p_i(y_i) + \epsilon \right)
    \]
    where $\epsilon$ is a small constant (e.g., $10^{-10}$) to avoid numerical instability when $p_i(y_i) \approx 0$.

    \item \textbf{Rank} of the true last token in the predicted token distribution:  
    The rank $r_i$ is defined as:
    \[
        r_i = 1 + \sum_{t \neq y_i} \mathbb{I} \left[ p_i(t) > p_i(y_i) \right]
    \]
    where $\mathbb{I}[\cdot]$ is the indicator function, counting the number of tokens with a higher predicted probability than the true token.
\end{enumerate}

This evaluation allows us to compare how neuron selection based on a global concept representation is effective.

\subsubsection{Bias}
For the bias evaluation, we use the \emph{Bias in Bios} dataset, which contains short biographical text snippets annotated with the gender of the person being described. From the test split, we randomly sample $400$ bios. Each bio is associated with a binary gender label: $0$ for male and $1$ for female.

We measure the effect of neuron deactivation on the model’s gender prediction ability using a \emph{resolution accuracy} metric. The evaluation is conducted as follows:
\begin{enumerate}
    \item We first evaluate the baseline resolution accuracy without deactivating any neurons.
    \item We then load the set of biased neurons (identified using absolute values of neuron effect scores, hereafter denoted as ``with abs'') for male and female concepts separately.
    \item For $k \in \{1, 3, 5, 7, 9, 11\}$, we deactivate:
    \begin{enumerate}
        \item The top-$k$ male-biased neurons.
        \item The top-$k$ female-biased neurons.
    \end{enumerate}
\end{enumerate}

Given a bio, the model is prompted with the text, and we inspect the probability distribution over the vocabulary for the next token. Let $\mathcal{M}$ and $\mathcal{F}$ denote the sets of masculine and feminine tokens, respectively from the model vocabulary. The masculine probability is computed as:
\[
    P_{\text{masc}} = \sum_{t \in \mathcal{M}} p(t)
\]
and the feminine probability as:
\[
    P_{\text{fem}} = \sum_{t \in \mathcal{F}} p(t)
\]
where $p(t)$ is the softmax-normalized probability assigned to token $t$.
The predicted gender $\hat{g}$ is determined as:
\[
    \hat{g} =
    \begin{cases}
        0 & \text{if } P_{\text{masc}} > P_{\text{fem}} \\
        1 & \text{otherwise}
    \end{cases}
\]
The \emph{resolution accuracy} for a gender $G$ is then:
\[
    \mathrm{ResAcc}(G) = \frac{\text{\# correctly predicted bios of gender } G}{\text{\# total bios of gender } G}
\]
We also compute:
\begin{itemize}
    \item \textbf{Average Resolution Accuracy}:
    \[
        \mathrm{ResAcc}_{\text{avg}} = \frac{\mathrm{ResAcc}(\text{male}) + \mathrm{ResAcc}(\text{female})}{2}
    \]
    \item \textbf{Gender Gap}:
    \[
        \mathrm{Gap} = \left| \mathrm{ResAcc}(\text{male}) - \mathrm{ResAcc}(\text{female}) \right|
    \]
\end{itemize}

This evaluation setup allows us to quantify both the predictive accuracy and the gender disparity of the model before and after targeted neuron deactivation based on biased neuron lists.

\subsubsection*{Bias Evaluation on the IndiBias Dataset}  
In addition to the \emph{Bias in Bios} evaluation, we also assess the effect of neuron deactivation on the \emph{IndiBias} dataset \cite{sahoo-etal-2024-indibias}, which contains sentence pairs differing only in their stereotypicality. Each pair consists of a \emph{stereotypical} sentence and an \emph{anti-stereotypical} counterpart, along with a label indicating which is which.

The evaluation procedure follows the same neuron deactivation settings described above: for $k \in \{1, 3, 5, 7, 9, 11\}$, we deactivate the top-$k$ male-biased neurons, the top-$k$ female-biased neurons, and both simultaneously.

For evaluating on this dataset, we measure the percentage of cases where the model assigns higher log-probability to the stereotypical sentence (\emph{Stereotype Preference \%}), to the anti-stereotypical sentence (\emph{Anti-Stereotype Preference \%}), and cases where both receive equal log-probability (\emph{Equal Preference \%}). Apart from assigning equal log probability, the ideal case would be to choose 50\% of the samples to be stereotypical and 50\% of the samples to be anti stereotypical. 

This evaluation allows us to quantify the extent to which targeted neuron deactivation shifts the model’s preference away from stereotypical completions and towards balanced or anti-stereotypical behavior in the context of Indian settings.

\subsection{Baselines}

We evaluate neuron importance using multiple scoring baselines.  First one is \textbf{(a)}, the method that \cite{yu-ananiadou-2024-neuron} proposes, i.e., \textbf{Log-probability increase and sentence-based}. It computes neuron importance for a given sentence $s$ by measuring the change in the log-probability of the top-predicted token $w^{*}$ upon ablation of neuron $(l, n)$:
\[
\Delta_{\log p}(l, n) = \log p(w^{*} \mid s) - \log p_{-(l, n)}(w^{*} \mid s).
\]
Then, the following baselines are as described. We follow the most baselines that previous works such as \cite{yu-ananiadou-2024-neuron} use. Let $m$ denote the activation coefficient of a neuron after the first feed-forward layer ($c_{\mathrm{fc}}$), $v \in \mathbb{R}^d$ the corresponding second-layer weight vector ($c_{\mathrm{proj}}$), and $\|v\|$ its $\ell_2$ norm. Let $w$ be the target token, $\text{rank}(w)$ its rank under the model's predicted distribution, and $p(w \mid \cdot)$ the predicted probability. We define the following baselines: \textbf{(b)} \textbf{Log-probability}: $\log p(w \mid m v)$, where $m v$ is the neuron's contribution to the residual stream; \textbf{(c)} \textbf{Probability difference}: $p(w \mid m v + A + h) - p(w \mid A + h)$, where $A$ and $h$ denote the contributions of attention and other hidden states, respectively; \textbf{(d)} \textbf{Weight norm}: $\|v\|$; \textbf{(e)} \textbf{Activation magnitude}: $|m|$; \textbf{(f)} \textbf{Activation--weight product}: $|m| \cdot \|v\|$; and \textbf{(g)} \textbf{Activation--rank product}: $\frac{|m|}{\text{rank}(w)}$. Each baseline is computed for all neurons across all layers, and the top-$n$ neurons are selected according to the respective scoring function.

\subsection{Analysis of Results}
In this section, we analyze the the results obtained following the evaluation pipeline discussed previously.
\subsubsection{Hate Speech evaluation}
\begin{table*}[htbp]
\centering
\caption{Average of the metrics across all sentences involved in testing for our method and baselines. Lower the metrics, better the performance. Denoted best performing method in bold. \textit{Baseline a through g are described in the Baselines section.}}
\begin{tabular}{lcccccccc}
\toprule
 & \textbf{Our Method} & \textbf{a} & \textbf{b} & \textbf{c} & \textbf{d} & \textbf{e} & \textbf{f} & \textbf{g} \\
\midrule
\textbf{$MRR_{avg.}$}  & \textbf{6.50e-05} & 0.102 & 1.0 & 1.0 & 1.0 & 0.180 & 0.95 & 0.180 \\
\textbf{$Prob_{avg.}$} & \textbf{5.28e-06} & 0.100 & 0.853 & 0.853 & 0.987 & 0.007 & 0.396 & 0.007 \\
\textbf{$LogP_{avg.}$} & \textbf{-12.196} & -10.876 & -0.188 & -0.188 & -0.013 & -5.233 & -1.14 & -5.233 \\
\bottomrule
\end{tabular}
\label{tab:our_vs_baselines}
\end{table*}

In this section, we evaluate our method in a next-token prediction setting for the hate speech concept. As shown in Table~\ref{tab:our_vs_baselines}, our concept-vector based neuron attribution approach consistently outperforms all other baselines (\textbf{a} through \textbf{g}) across all metrics: Mean Reciprocal Rank ($MRR_{avg.}$), Probability ($Prob_{avg.}$), and Log-Probability ($LogP_{avg.}$). 

Specifically, deactivating the top 10 neurons identified by our method causes a dramatic performance degradation, with $MRR_{avg.} = 6.50 \times 10^{-5}$, $Prob_{avg.} = 5.28 \times 10^{-6}$, and $LogP_{avg.} = -12.196$. In comparison, the best-performing baseline (\textbf{a}) has $MRR_{avg.} = 0.102$ and $Prob_{avg.} = 0.100$, which are orders of magnitude higher, indicating that the neurons it identifies are much less impactful for this concept.

The magnitude of this drop implies that the neurons discovered by our method are strongly associated with the representation of hate speech, and their removal effectively eliminates the model’s ability to produce the correct final token in hateful sentences. Since lower metric values indicate greater disruption, this demonstrates the effectiveness of our approach in identifying concept-relevant neurons.

Furthermore, Table~\ref{tab:baseline_b_metrics} highlights that neurons obtained on a per-example basis (i.e., from one sentence) fail to generalize when evaluated on other sentences. For instance, baseline~\textbf{b} obtains perfect $MRR = 1.0$ on its source sentence (0) but fails to reduce the probability meaningfully when tested on a different sentence (2), maintaining $MRR = 1.0$ and $Prob \approx 1.0$. In contrast, our method’s neurons—derived from a global concept vector—consistently yield low scores across all test sentences, as shown in Table~\ref{tab:sentence_metrics_0_9}, where $MRR$ values range between $5.27 \times 10^{-5}$ and $9.40 \times 10^{-5}$, $Prob$ remains in the $10^{-6}$ range, and $LogP$ is consistently around $-12$. This stability confirms that our method generalizes across examples rather than overfitting to individual instances.

In the tables, only indices of sentences are shown, because the examples are related to hate and are not ethical to be included in the manuscript. We plan to release the exact sentences as supplementary material.

\begin{table}[htbp]
\centering
\caption{Metrics for neurons obtained using different baseline methods. From set is the index of the sentence from which important neurons are obtained and Test sent is the sentence which is tested by turning off the neurons obtained. We can observe that the neurons obtained through baseline methods on per-example basis do not perform consistently well across all the sentences. \textit{(Baselines a through g are described in the Baselines section.)}}
\resizebox{\columnwidth}{!}{%
\begin{tabular}{lcccccc}
\toprule
\textbf{Baseline} & \textbf{From Sent} & \textbf{Test Sent} & \textbf{MRR} & \textbf{Prob} & \textbf{LogP} \\
\midrule
a & 0 & 0 & 2.91e-05 & 4.87e-07 & -14.5354 \\
 & 0 & 2 & 2.91e-05 & 5.45e-07 & -14.4218 \\
 & 2 & 0 & 9.32e-04 & 1.54e-04 & -8.7808 \\
b & 0 & 0 & 1.0000 & 0.4532 & -0.7914 \\
 & 0 & 2 & 1.0000 & 0.9967 & -0.0033 \\
 & 2 & 0 & 1.0000 & 0.5688 & -0.5642 \\
c & 0 & 0 & 1.0000 & 0.4532 & -0.7914 \\
 & 0 & 2 & 1.0000 & 0.9967 & -0.0033 \\
 & 2 & 0 & 1.0000 & 0.5688 & -0.5642 \\
d & 0 & 0 & 1.0000 & 0.9126 & -0.0915 \\
 & 0 & 2 & 1.0000 & 0.9999 & -7.01e-05 \\
 & 2 & 0 & 1.0000 & 0.9126 & -0.0915 \\
e & 0 & 0 & 0.0089 & 0.0011 & -6.8421 \\
 & 0 & 2 & 0.0714 & 0.0049 & -5.3230 \\
 & 2 & 0 & 0.0714 & 0.0049 & -5.3230 \\
f & 0 & 0 & 1.0000 & 0.0955 & -2.3485 \\
 & 0 & 2 & 1.0000 & 0.4328 & -0.8374 \\
 & 2 & 0 & 1.0000 & 0.0797 & -2.5291 \\
g & 0 & 0 & 0.0089 & 0.0011 & -6.8421 \\
 & 0 & 2 & 0.0714 & 0.0049 & -5.3230 \\
 & 2 & 0 & 0.0089 & 0.0011 & -6.8421 \\
\bottomrule
\end{tabular}%
}
\label{tab:baseline_b_metrics}
\end{table}

\begin{table}[htbp]
\centering
\caption{Metrics for each of the sentence, where neurons obtained through concept vector all at once. We can observe that the neurons obtained through concept vector at once performs consistently well across all the sentences.}
\begin{tabular}{lccc}
\toprule
\textbf{Sentence} & \textbf{MRR} & \textbf{Prob} & \textbf{LogP} \\
\midrule
0 & 9.40e-05 & 9.58e-06 & -11.5556 \\
1 & 7.82e-05 & 7.02e-06 & -11.8673 \\
2 & 5.60e-05 & 3.92e-06 & -12.4497 \\
3 & 5.27e-05 & 3.38e-06 & -12.5983 \\
4 & 8.13e-05 & 7.75e-06 & -11.7673 \\
5 & 6.75e-05 & 5.29e-06 & -12.1488 \\
6 & 5.84e-05 & 4.29e-06 & -12.3597 \\
7 & 8.57e-05 & 8.34e-06 & -11.6939 \\
8 & 6.41e-05 & 5.39e-06 & -12.1312 \\
9 & 5.99e-05 & 4.59e-06 & -12.2924 \\
\bottomrule
\end{tabular}
\label{tab:sentence_metrics_0_9}
\end{table}

\subsubsection{Bias evaluation}
We next evaluate our method in a gender bias resolution task, where the goal is to locate neurons associated with predicting the correct gendered resolution (male or female) in ambiguous contexts. Using our concept vector formulation, we identify separate sets of male and female neurons and then progressively deactivate them to test whether they are indeed responsible for gender classification.

Table~\ref{tab:male_neurons} shows that deactivating male neurons results in a sharp decline in male accuracy (from $98.37\%$ in the original setting to just $5.00\%$ after removing 11 neurons), while female accuracy increases correspondingly (from $95.08\%$ to $99.21\%$). The gender gap also increases dramatically—from $0.0328$ in the original model to $0.9421$ with 11 neurons deactivated—indicating a strong suppression of male-specific representation.

Conversely, Table~\ref{tab:female_neurons} exhibits the reverse effect: female accuracy plummets from $95.08\%$ to just $2.90\%$ after deactivating 11 neurons, while male accuracy remains above $96\%$. The gender gap likewise jumps to $0.9363$, confirming that these neurons are indeed responsible for representing female-specific information.

These results validate that our method successfully identifies neurons tied to high-level conceptual features—in this case, gender resolution—and that their selective deactivation yields predictable and interpretable changes in model behavior.

\begin{table}[h]
\centering
\caption{Effect of Deactivating Male Neurons - We can observe decrease in the male resolution accuracy with increase in number of male neurons we turn off}
\label{tab:male_neurons}
\begin{tabular}{@{}ccccc@{}}
\toprule
\textbf{\# Neurons} & \textbf{Male Acc.} & \textbf{Female Acc.} & \textbf{Avg. Acc.} & \textbf{Gender Gap} \\ \midrule
Original  & 0.9837 & 0.9508 & 0.9672 & 0.0328 \\
1  & 0.6920 & 0.9856 & 0.8388 & 0.2937 \\
3  & 0.1185 & 0.9909 & 0.5547 & 0.8724 \\
5  & 0.0752 & 0.9914 & 0.5333 & 0.9162 \\
7  & 0.0627 & 0.9917 & 0.5272 & 0.9290 \\
9  & 0.0543 & 0.9919 & 0.5231 & 0.9376 \\
11 & 0.0500 & 0.9921 & 0.5210 & 0.9421 \\ \bottomrule
\end{tabular}
\end{table}

\begin{table}[h]
\centering
\caption{Effect of Deactivating Female Neurons - We can observe decrease in the female resolution accuracy with increase in number of female neurons we turn off}
\label{tab:female_neurons}
\begin{tabular}{@{}ccccc@{}}
\toprule
\textbf{\# Neurons} & \textbf{Male Acc.} & \textbf{Female Acc.} & \textbf{Avg. Acc.} & \textbf{Gender Gap} \\ \midrule
Original  & 0.9837 & 0.9508 & 0.9672 & 0.0328 \\
1  & 0.9764 & 0.1393 & 0.5578 & 0.8372 \\
3  & 0.9615 & 0.0431 & 0.5023 & 0.9184 \\
5  & 0.9633 & 0.0362 & 0.4997 & 0.9270 \\
7  & 0.9654 & 0.0295 & 0.4975 & 0.9359 \\
9  & 0.9648 & 0.0292 & 0.4970 & 0.9356 \\
11 & 0.9652 & 0.0290 & 0.4971 & 0.9363 \\ \bottomrule
\end{tabular}
\end{table}

\subsubsection{IndiBias Evaluation}
Finally, we assess our method on the \textit{Indian Stereotypical Bias} task. An unbiased model should select stereotypical (S) and anti-stereotypical (A) completions with equal probability ($\approx 50\%$ each). However, in the original setting (Table~\ref{tab:Indi_Bias}), the model shows a clear bias: $S\% = 61.3$ and $A\% = 38.7$, a deviation of $+11.3$ percentage points from ideal balance.

We use our method to extract neurons representing male and female stereotypes separately, and then progressively deactivate them. For male stereotype neurons, $S\%$ drops steadily from $61.3$ (original) to $50.3$ after 11 neurons are removed, bringing the preference within $0.3$ percentage points of balance. Similar patterns emerge for female stereotype neurons, with $S\%$ decreasing to $50.1$ after deactivating 11 neurons.

The most balanced state occurs when we deactivate both one male and one female stereotype neuron simultaneously, yielding $S\% = 50.6$ and $A\% = 49.4$. This is within $0.6$ percentage points of perfect parity, demonstrating that the neurons identified by our method effectively capture the stereotypical bias and that their removal neutralizes it.

Overall, across all three evaluations, the numerical trends confirm that our concept-vector based neuron attribution consistently finds neurons that represent broad, generalizable concepts, rather than idiosyncratic features tied to specific examples.

\begin{table}[ht]
\centering
\caption{Effect of neuron deactivation on stereotype preference. S-Cnt denotes the number of times the model is preferring Stereotypical example over an Anti-Stereotypical example and A-Cnt denotes the number of times the model is preferring an Anti-Stereotypical example over a Stereotypical example. Similarly, S\% and A\% denotes the percentage of the times the model preferring Stereotypical example and Anti-Stereotypical example respectively.}
\label{tab:Indi_Bias}
\resizebox{\columnwidth}{!}{
\begin{tabular}{lrrrrrr}
\toprule
\textbf{Setting} & \textbf{Total} & \textbf{S-Cnt} & \textbf{A-Cnt} & \textbf{S\%} & \textbf{A\%}\\
\midrule
No Deactivation & 561 & 344 & 217 & 61.3 & 38.7 \\
1 Male Neuron & 561 & 311 & 250 & 55.4 & 44.6 \\
3 Male Neurons & 561 & 314 & 247 & 56.0 & 44.0 \\
5 Male Neurons & 561 & 293 & 268 & 52.2 & 47.8 \\
7 Male Neurons & 561 & 289 & 272 & 51.5 & 48.5 \\
9 Male Neurons & 561 & 289 & 272 & 51.5 & 48.5 \\
11 Male Neurons & 561 & 282 & 279 & 50.3 & 49.7 \\
1 Female Neuron & 561 & 320 & 241 & 57.0 & 43.0 \\
3 Female Neurons & 561 & 294 & 267 & 52.4 & 47.6 \\
5 Female Neurons & 561 & 304 & 257 & 54.2 & 45.8 \\
7 Female Neurons & 561 & 286 & 275 & 51.0 & 49.0 \\
9 Female Neurons & 561 & 286 & 275 & 51.0 & 49.0 \\
11 Female Neurons & 561 & 281 & 280 & 50.1 & 49.9 \\
1 Male + 1 Female & 561 & 284 & 277 & 50.6 & 49.4 \\
\bottomrule
\end{tabular}
}
\end{table}

From all the cases above, and the results we observe, we can say that our method is successful in finding neurons that are responsible for a particular broad level concepts and does not rely on per-example basis.
\section{Conclusion}
We proposed \textit{NEAT}, a concept-vector-based method for efficiently locating neurons in large language models that encode human-interpretable concepts. By representing a concept as a single dense vector, our approach reduces the forward-pass complexity from $\mathcal{O}(n \times m)$ to $\mathcal{O}(n)$, enabling scalable neuron attribution without sacrificing interpretability.

Across hate speech, gender bias, and Indian stereotypical bias evaluations, \textit{NEAT} consistently identified neurons whose targeted removal caused significant and predictable shifts in model behavior. In hate speech detection, deactivating the top-10 neurons led to orders-of-magnitude drops in MRR, probability, and log-probability compared to baselines. In bias evaluations, removing gender-specific neurons systematically altered model predictions, and in the Indian bias setting, yielded near-balanced stereotype preferences.

Unlike per-example methods, \textit{NEAT} generalizes across inputs, capturing concept-level representations that are robust and consistent. These results highlight its potential for targeted, interpretable interventions, and open avenues for future work on multi-concept analysis, dynamic neuron control, and integration with model editing for safer, more aligned LLMs.
\bibliographystyle{ACM-Reference-Format}
\bibliography{main}

\appendix
\section{Appendix}
\label{sec:appendix}
\begin{table}[h]
\centering
\scriptsize
\caption{Resolution Accuracy and Gender Gap for Different Clustering Configurations (KMeans on Coefficient Scores).}
\label{tab:clustering_results}
\begin{tabular}{lcccc}
\hline
\textbf{Clustering Setting} & \textbf{Male Acc.} & \textbf{Female Acc.} & \textbf{Average Acc.} & \textbf{Gender Gap} \\
\hline
No Neuron Deactivation & 0.9837 & 0.9508 & \textbf{0.9672} & 0.0328 \\
Cluster of 2 (Top Cluster) & 0.8792 & 0.5230 & 0.7012 & 0.3562 \\
Cluster of 2 (Other 20 Clusters) & 0.3814 & 0.1863 & 0.2839 & 0.1951 \\
Cluster of 5 & 0.7018 & 0.4682 & 0.5850 & 0.2336 \\
Cluster of 10 & 0.6367 & 0.4081 & 0.5224 & 0.2286 \\
Cluster of 20 & 0.5183 & 0.2941 & 0.4062 & 0.2242 \\
Cluster of 33 & 0.4269 & 0.4053 & 0.4161 & \textbf{0.0216} \\
Cluster of 50 & 0.3584 & 0.1928 & 0.2756 & 0.1656 \\
Cluster of 75 & 0.2216 & 0.1721 & 0.1969 & 0.0495 \\
Cluster of 100 & 0.1682 & 0.0974 & 0.1328 & 0.0708 \\

\hline
\end{tabular}
\end{table}

\paragraph{Analysis.}  
The results reveal that the highest resolution accuracy is achieved when no neuron is deactivated. Once neurons begin to be deactivated, the resolution accuracy steadily declines, indicating that these neurons are not merely responsible for storing gender bias, but rather for associating gender with gendered pronouns present in the dataset. Removing them therefore disrupts the model’s ability to correctly resolve pronouns.  

Furthermore, the gender gap initially increases with neuron deactivation, suggesting that neurons associated with female gendered pronouns are removed earlier. The gap then reaches its minimum when 33 neurons are deactivated, implying that at this point, neurons linked to both male and female gendered pronouns have been removed, resulting in a more balanced—though overall less accurate—model performance.


\end{document}